\documentclass[11pt]{article}

\usepackage[preprint]{acl}

\usepackage{times}
\usepackage{latexsym}

\usepackage[T1]{fontenc}

\usepackage[utf8]{inputenc}

\usepackage{microtype}
\usepackage{inconsolata}
\usepackage{graphicx}
\usepackage{subfigure} 
\usepackage{amsmath}
\usepackage{stfloats}
\usepackage{multirow}
\usepackage{enumitem}
\usepackage{booktabs}
\usepackage{bbding}
\usepackage{amsfonts,amssymb}
\usepackage{diagbox}
\usepackage{newfloat}
\usepackage{listings}
\usepackage{times}  
\usepackage{helvet} 
\usepackage{courier}  
\usepackage{natbib}  
\usepackage{caption}
\usepackage{makecell}
\usepackage{xcolor}
\usepackage{pifont}
\usepackage{soul}
\usepackage{url}
\usepackage{tikz}
\usepackage{colortbl}
\usepackage{xspace}
\newcommand{\tabincell}[2]{\begin{tabular}{@{}#1@{}}#2\end{tabular}}

\definecolor{best}{RGB}{202,252,209}
\definecolor{second}{RGB}{250, 229, 215}
\definecolor{mention}{RGB}{255, 242, 206} 
\definecolor{mypink}{HTML}{d896b3}
\definecolor{myorange}{HTML}{ff8d2b}
\definecolor{myblue}{HTML}{1f59a1}
\definecolor{mygreen}{HTML}{3b93bb}
\definecolor{mygrey}{HTML}{aaaa9e}
\definecolor{myyellow}{HTML}{f0c369}
\definecolor{mypurple}{HTML}{7f439c}
\definecolor{myspike}{HTML}{1e9695}
\definecolor{gaincolor}{HTML}{228B22}

\newcommand{\model}{\textbf{\textsc{FigR}}\xspace}

\title{
\raisebox{-0.3\height}{%
  \includegraphics[width=0.8cm]{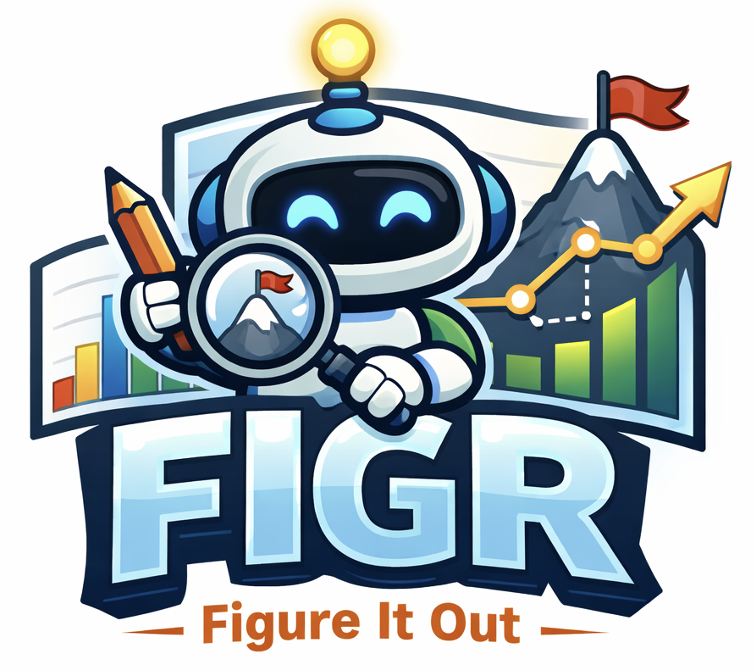}
}
\textit{Figure} It Out: Improve the Frontier of Reasoning with \\
Executable Visual States
}

\author{
 Meiqi~Chen, 
Fandong~Meng\footnotemark[2],
 Jie~Zhou \\ 
 WeChat AI, Tencent Inc \\
 \texttt{\{meiqiichen, fandongmeng, withtomzhou\}@tencent.com} 
}

\begin{document}

\makeatletter
\def\@maketitle{\vbox to \titlebox{\hsize\textwidth
 \linewidth\hsize \vskip 0.125in minus 0.125in \centering
 {\Large\bfseries \@title \par}
 \vskip 0.1in
 {\def\and{\unskip\enspace{\rmfamily and}\enspace}%
  \def\And{\end{tabular}\hss \egroup \hskip 1in plus 2fil
           \hbox to 0pt\bgroup\hss \begin{tabular}[t]{c}\bfseries}%
  \def\AND{\end{tabular}\hss\egroup \hfil\hfil\egroup
          \vskip 0.25in plus 1fil minus 0.125in
           \hbox to \linewidth\bgroup\large \hfil\hfil
             \hbox to 0pt\bgroup\hss \begin{tabular}[t]{c}\bfseries}
  \hbox to \linewidth\bgroup\large \hfil\hfil
    \hbox to 0pt\bgroup\hss
  \outauthor
   \hss\egroup
    \hfil\hfil\egroup}

  \vskip 0.1in
{\centering
   \includegraphics[width=0.97\textwidth,
    clip,trim=0pt 0pt 5pt 0pt]{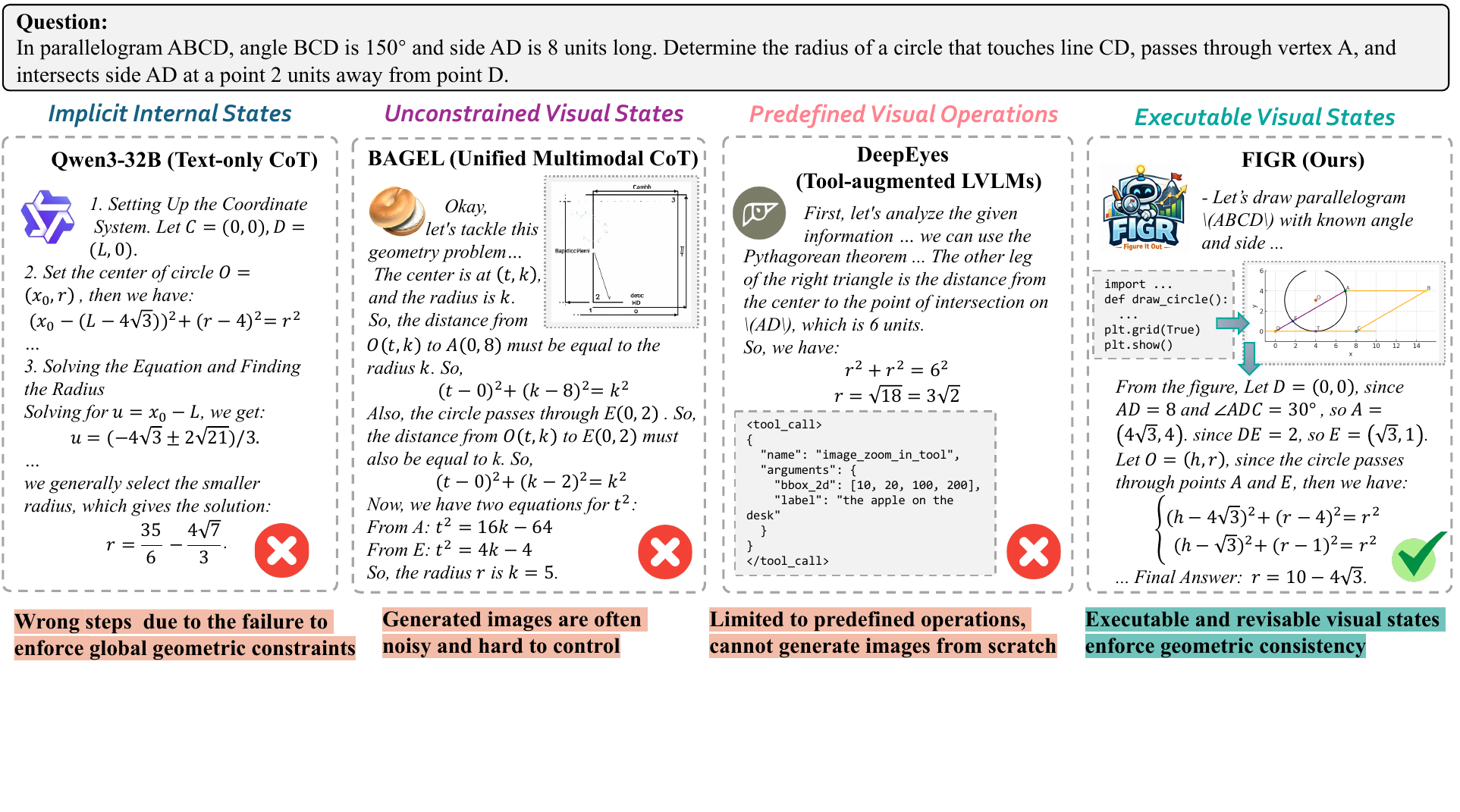}\par}
  \vspace{-3.3em}

\captionsetup{type=figure}
{\small
 \noindent\hspace{0.02\textwidth}%
 \parbox{0.96\textwidth}{%
   \refstepcounter{figure}%
   \label{fig:example}%
   \caption*{\figurename~\thefigure: 
Comparison of reasoning paradigms by intermediate state representation.
Text-only reasoning relies on implicit symbolic states, unified multimodal models generate uncontrolled visual states, and tool-augmented models operate on given images with predefined transformations.
In contrast, \model constructs executable visual states within the reasoning loop, enabling precise, revisable diagrams that enforce geometric consistency.

}%
 }%
}

}}
\makeatother

\maketitle
\begin{abstract}
Complex reasoning problems often involve implicit spatial and geometric relationships that are not explicitly encoded in text. While recent reasoning models perform well across many domains, purely text-based reasoning struggles to capture structural constraints in complex settings.
In this paper, we introduce \model, which integrates executable visual construction into multi-turn reasoning via end-to-end reinforcement learning. Rather than relying solely on textual chains of thought, \model externalizes intermediate hypotheses by generating executable code that constructs diagrams within the reasoning loop. An adaptive reward mechanism selectively regulates when visual construction is invoked, enabling more consistent reasoning over latent global properties that are difficult to infer from text alone.
Experiments on eight challenging mathematical benchmarks demonstrate that \model outperforms strong text-only chain-of-thought baselines, improving the base model by 13.12\% on AIME 2025 and 11.00\% on BeyondAIME. These results highlight the effectiveness of precise, controllable figure construction of \model in enhancing complex reasoning ability. 
\end{abstract}

\renewcommand*{\thefootnote}{\fnsymbol{footnote}}

\footnotetext[1]{\small \url{https://github.com/chenmeiqii/FIGR}.}
\footnotetext[2]{Corresponding author.}

\section{Introduction}
Human reasoning is deeply intertwined with the ability to externalize information into visual form. When faced with complex constraints or multi-step relationships, people naturally draw diagrams, sketch intermediate states, or construct spatial layouts to reduce cognitive load and clarify underlying structure \cite{tversky2019diagrams, norman2024things}. 
Cognitive science further suggests that such diagrammatic reasoning allows people to discover patterns that are difficult or even impossible to glean from text alone \cite{donald1993origins, mandler2010spatial}. 

Figure~\ref{fig:example} provides an illustrative comparison of how different reasoning paradigms behave on a geometry problem involving angle constraints, tangency conditions, and point intersections.
In text-only chain-of-thought (CoT, \cite{wei2022chain}) models, all spatial relations must be implicitly maintained through symbolic expressions.  This places a heavy burden on internal representations and often leads to cascading algebraic errors when subtle geometric constraints are misinterpreted.
A natural extension is to consider unified multimodal models ~\cite{team2024chameleon, wu2025janus, deng2025emerging, li2025zebra}, which generate images as part of the reasoning process to make spatial relationships explicit. While conceptually appealing, these models lack precise control over the generated visual content. Since image generation is not grounded in executable constraints, even small spatial inconsistencies can propagate across reasoning steps, limiting their reliability in problems that require fine-grained geometric precision.

In parallel, tool-augmented large vision-language models (LVLMs) incorporate external visual tools or predefined APIs to assist reasoning, spanning prompt engineering \cite{gupta2023visual, hu2024visual, suris2023vipergpt}, fine-tuning \cite{Wu_2024_CVPR, wang2025pixel}, or reinforcement-learning \cite{zheng2025deepeyes, zhang2025chain}. By delegating visual operations to explicit tools, these approaches improve controllability and execution fidelity.
However, they are fundamentally constrained to operating on \emph{given} images and predefined transformations (e.g., zooming or cropping), which prevents them from autonomously constructing task-specific diagrams required by complex reasoning problems.

Motivated by these limitations, we introduce \model, which integrates visual construction directly into the reasoning loop via end-to-end reinforcement learning.
Rather than relying on unconstrained image generation or fixed toolsets, \model produces executable code that bridges symbolic reasoning and visual rendering.
This allows the model to actively construct and iteratively refine diagrams during multi-turn inference, using rendered figures as explicit, stateful feedback---analogous to how humans repeatedly sketch and update diagrams when solving complex problems. As illustrated in Figure~\ref{fig:example}, this mechanism enforces geometric consistency and supports correct reasoning in cases where other paradigms fail.

To regulate this reasoning–rendering process, we design an \emph{Adaptive Reward Mechanism} that selectively encourages visual construction when it is beneficial, while discouraging unnecessary or spurious rendering behaviors. Importantly, this design removes the need for a supervised visual reasoning cold-start stage. Instead, \model can be initialized from an instruction-tuned model (e.g., \texttt{Qwen3-VL-32B-Instruct} \cite{bai2025qwen3vl}) and refined purely through reinforcement learning. Through this process, \model autonomously learns both when to invoke visual reasoning and how to integrate execution feedback into its reasoning trajectory, without relying on task-specific supervised visual reasoning data.

Overall, our main contributions are as follows: 
\begin{itemize}[leftmargin=*, itemsep=0pt, parsep=0pt]
\item We propose \model, which actively conducts visual thinking during multi-turn reasoning.
\item We introduce an adaptive reward mechanism that regulates the selective use of visual reasoning without requiring supervised cold-start.
\item We demonstrate substantial improvements over strong text-only baselines across diverse mathematical reasoning benchmarks.
\end{itemize}
\section{Related Work}
\subsection{Text-based and Programmatic Reasoning}
A series of works extends chain-of-thought (CoT) prompting \cite{wei2022chain} by introducing structured reasoning strategies for large language models (LLMs) \cite{achiam2023gpt, touvron2023llama}. Program-of-Thought (PoT) \cite{chen2022program} and Chain-of-Code \cite{li2023chain} interleave natural language with executable code, allowing models to offload arithmetic and symbolic manipulation to external interpreters. Tree-of-Thought (ToT) \cite{yao2023tree} and its variants further represent reasoning as a branching search process, enabling exploration of alternative solution paths and pruning of unpromising branches. More recently, reinforcement learning has been applied to regulate reasoning behaviors, such as learning when to expand or truncate reasoning trajectories \cite{guo2025deepseek}, or when and how to invoke a code interpreter during problem solving \cite{feng2025retool}.

Despite these advances, such approaches operate primarily over textual or symbolic representations. Intermediate reasoning states remain implicit and must be maintained internally by the model, which poses fundamental challenges for tasks involving geometry, kinematics, or other forms of spatial reasoning. Without explicit external representations, complex relational constraints are prone to the accumulation of errors, limiting the reliability of purely text-based and programmatic reasoning in structurally demanding settings.

\subsection{Unconstrained Visual Generation for Reasoning}
Large vision-language models (LVLMs) \cite{liu2023visual, bai2025qwen2} demonstrate that LLMs can interpret images through relatively shallow alignment layers, but most of them remain limited to text-only outputs. Recent unified multimodal architectures \cite{team2024chameleon, wu2025janus, deng2025emerging, li2025zebra} further extend generation to both text and images within a single model, enabling visual content to be produced as part of the reasoning process.

While conceptually appealing, these approaches rely on unconstrained image generation without explicit mechanisms to enforce geometric precision or relational consistency. As a result, the generated visual representations are often noisy or spatially imprecise, limiting their effectiveness as reliable intermediate states for reasoning. This limitation becomes particularly pronounced in mathematical and scientific domains, where even minor visual inaccuracies can propagate across reasoning steps and lead to significant downstream errors.

\subsection{Tool-Augmented Visual Reasoning}
Another line of research augments LVLMs with external visual tools, rendering engines, or predefined APIs. These methods decompose a problem into sub-steps, apply visual operators such as zooming, cropping, or drawing auxiliary lines, and interpret the results to guide subsequent reasoning \cite{gupta2023visual, suris2023vipergpt, hu2024visual, shao2024visual, Wu_2024_CVPR, li2025dyfo, shen2025zoomeye, wang2025pixel}. Reinforcement learning-based pipelines further improve tool usage efficiency by explicitly rewarding appropriate invocation patterns \cite{zheng2025deepeyes, zhang2025chain}.

Tool-augmented visual reasoning approaches provide strong controllability, as their visual operations are deterministic and explicitly defined. However, they are inherently restricted to operating on given images and a fixed set of predefined transformations. Consequently, they lack the ability to autonomously construct new diagrams, geometric configurations, or abstract visual representations that are not present in the input. This restriction limits their applicability in reasoning tasks that require dynamic generation and refinement of task-specific visual structures.

\section{Methodology}
\begin{figure*}[t]
\centering  
\includegraphics[width=0.95\textwidth]{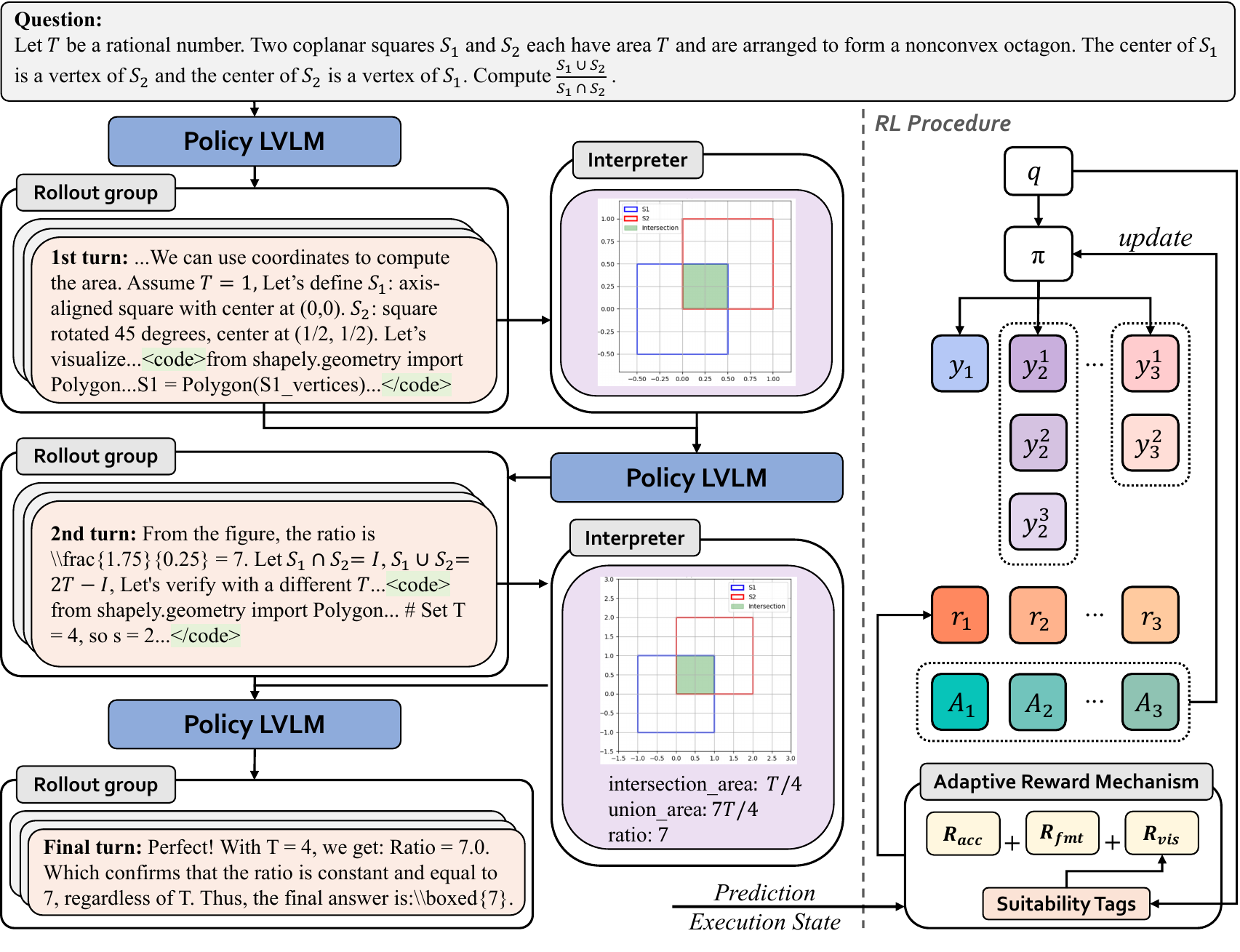}
\caption{Overview of \model. \model alternates between textual reasoning and \emph{executable visual construction} within a unified reasoning loop. An adaptive reward mechanism selectively regulates when visual construction is invoked, based on question suitability, execution outcomes, and final answer correctness.}
\label{fig:method}
\vspace{-4mm}
\end{figure*}

In this section, we describe the main components of \model. 
Section~\ref{subsec:overview} introduces the executable reasoning loop, Section~\ref{subsec:rl} details the reinforcement learning procedure, and Section~\ref{subsec:arm} presents the adaptive reward mechanism.

\subsection{Executable Visual Construction within the Reasoning Loop}
\label{subsec:overview}
\model integrates executable visual construction directly into its multi-turn reasoning process. For each input, the model interleaves textual inference and code generation to construct diagrams, analogous to how humans iteratively sketch intermediate states while reasoning.

Formally, at each reasoning step $t$, the model maintains a state consisting of a textual context $h_t$ (including prior reasoning steps, code outputs, and interpreter feedback) and an optional rendered diagram $I_t$. The policy $\pi_\theta$ samples an action:

\begin{equation}
    a_t \sim \pi_\theta\left(\cdot \mid h_t, I_t\right) ,
\end{equation}
where $a_t$ is either (i) a textual continuation or (ii) a code snippet $c_t$ generated by the policy itself.
When a code action is emitted, it is executed by a sandboxed interpreter, which may produce textual feedback $T_{t+1}$ and/or a rendered diagram $I_{t+1}$:

\begin{equation}
\big(T_{t+1}, I_{t+1}\big)=\operatorname{Interpreter}\left(c_t\right),
\end{equation}

The context is then updated accordingly:

\begin{equation}
h_{t+1}=\operatorname{UpdateContext}\left(h_t, c_t, T_{t+1}, I_{t+1}\right) .
\end{equation}

This process repeats until a termination token is emitted or a step limit is reached. By representing diagrams as executable and reproducible states, \model enables precise and controllable visual feedback throughout the reasoning process.

\subsection{Reinforcement Learning with Multi-turn Executable Reasoning}
\label{subsec:rl}
Training \model requires learning long-horizon reasoning policies that interleave textual inference with executable visual construction. In this setting, the quality of individual actions (e.g., whether to emit code at a given step) cannot be evaluated locally, as their utility depends on the final reasoning outcome and the resulting visual states. We therefore formulate learning at the trajectory level and employ reinforcement learning to optimize multi-turn reasoning behaviors.

We adopt Group Relative Policy Optimization (GRPO) \cite{shao2024deepseekmath}, which is well-suited for trajectory-level supervision with delayed terminal rewards. For each input question, GRPO samples a group of $G$ candidate trajectories under the current policy. Each trajectory $\tau_i$ consists of a sequence of textual and code actions, together with their executed interpreter feedback, and terminates when the model emits a special end token.
After rollout termination, we compute a scalar reward:
\begin{equation}
R_i = R_{\mathrm{acc}}(\tau_i) + R_{\mathrm{fmt}}(\tau_i) + R_{\mathrm{vis}}(\tau_i),
\end{equation}
which jointly reflects final answer correctness, output format compliance, and the appropriateness of visual construction (Section~\ref{subsec:arm}). Importantly, rewards are assigned only at the trajectory level, as intermediate actions contribute to reasoning quality through their cumulative effects.

To stabilize learning under such delayed supervision, GRPO compares trajectories generated from the same prompt and uses the group-average reward as a baseline:
\begin{equation}
\bar{R}=\frac{1}{G} \sum_{i=1}^G R_i.
\end{equation}

The relative advantage of each trajectory is then defined as:
\begin{equation}
\hat{A}_i=R_i-\bar{R},
\end{equation}

which provides a low-variance learning signal that emphasizes relative behavioral differences, such as when executable visual construction leads to better outcomes than purely textual reasoning.

The policy is updated by maximizing the following surrogate objective:
\begin{equation}
\small
\begin{aligned}
J_{\mathrm{GRPO}}(\theta)
= &\ \mathbb{E}_{q\sim\mathcal{D},\{\tau_i\}_{i=1}^G\sim\pi_{\theta_{\text{old}}}} \frac{1}{G}\sum_{i=1}^G \frac{1}{|\tau_i|}
\sum_{t=1}^{|\tau_i|} M_{i,t}(\theta)
\\
& - \beta\, D_{\mathrm{KL}}(\pi_\theta \,\|\, \pi_{\mathrm{ref}}),
\end{aligned}
\end{equation}
where
\begin{equation}
\small
\begin{aligned}
M_{i,t}(\theta)
= \min\!\left[
r_{i,t}(\theta)\hat A_i,\ 
\operatorname{clip}(r_{i,t}(\theta),1-\varepsilon,1+\varepsilon)\hat A_i
\right].
\\
\end{aligned}
\end{equation}

Here, $r_{i,t}(\theta)=\frac{\pi_\theta(a_{i,t} \mid s_{i,t})}{\pi_{\theta_{\text{old}}}(a_{i,t} \mid s_{i,t})}$ denotes the probability ratio between the updated and previous policies, $\varepsilon$ is the clipping threshold, and $\pi_{\mathrm{ref}}$ is a reference policy used for KL regularization. This objective encourages the policy to increase the likelihood of trajectories that achieve higher relative rewards, while maintaining stability during optimization.

Overall, GRPO enables \model to learn multi-turn reasoning strategies in which the decision to invoke visual construction and the resulting reasoning quality are optimized jointly, without requiring step-wise supervision or an explicit value model.

\subsection{Adaptive Reward Mechanism}
\label{subsec:arm}
While executable visual construction can significantly benefit certain reasoning problems, indiscriminate invocation of diagram generation may introduce unnecessary complexity or distract from effective textual reasoning. To address this trade-off, we introduce an adaptive reward mechanism that acts as a \emph{policy-level control signal}, guiding when visual construction should be invoked during multi-turn reasoning.

For each input question, we first estimate whether diagrammatic reasoning is likely to be beneficial by querying an auxiliary language model classifier, which outputs a binary suitability label $s \in \{0,1\}$. Importantly, this classifier does not provide step-wise supervision; instead, it modulates reward magnitude, encouraging selective use of visual construction and making the policy robust to moderate misclassification.

During reinforcement learning, the visual-invocation reward $R_{\mathrm{vis}}$ is evaluated only at the trajectory level and is coupled with final answer correctness. Specifically, a positive reward is granted when executable visual construction contributes to a correct solution, with higher weight assigned when diagrammatic reasoning is predicted to be appropriate for the task:

\begin{equation}
R_{\mathrm{vis}} =
\begin{cases}
1.0, & \text{if } y=\hat{y},\ s=1,\ \mathrm{exec}=1, \\
0.2, & \text{if } y=\hat{y},\ s=0,\ \mathrm{exec}=1, \\
0, & \text{otherwise},
\end{cases}
\end{equation}

Here, $y$ and $\hat{y}$ denote the ground-truth and predicted final answers, respectively, and $\mathrm{exec}$ indicates successful execution of the generated code.

Rather than explicitly supervising individual drawing actions, this reward formulation biases the policy toward reasoning strategies in which visual construction is used purposefully and effectively. By tying visual rewards to overall task success, the model learns to treat diagram construction as a meaningful intermediate state rather than a mandatory or habitual operation. This design discourages redundant or spurious visual actions while preserving the flexibility to exploit executable visual feedback when it improves reasoning outcomes.
\section{Experiments}
\subsection{Experimental Setup}
\label{subsec:setup}
\paragraph{Implementation Details}  
We conduct reinforcement learning using the VeRL framework \cite{sheng2024hybridflow}, with \texttt{Qwen3-VL-32B-Instruct} \cite{bai2025qwen3vl} as the base policy. We set the maximum number of interaction rounds to 3, the maximum generation length to 32{,}768 tokens, and the sampling temperature to 0.7.

\paragraph{Training Dataset}
We train \model on DeepMath-103K \cite{he2025deepmath}, a rigorously decontaminated dataset of 103{,}000 challenging mathematical problems with verifiable answers. Following Section~\ref{subsec:arm}, we annotate each instance with a binary suitability label indicating whether executable visual construction is likely to be beneficial. An auxiliary language model (Deepseek-V3 \cite{liu2024deepseek}) is used for this annotation; prompt details are provided in Appendix~\ref{subsec:prompt_classifier}.

\paragraph{Evaluation Datasets}
To ensure comparability with prior work, we conduct evaluations on a suite of mathematical reasoning datasets that are widely used as standard benchmarks. We use {AIME 2024 \cite{aime_2024}}, {AIME 2025 \cite{aime_2025}}, {BeyondAIME \cite{seed2025seed1}}, {MATH 500 \cite{hendrycks2021measuring, lightman2023let}}, {AMC \cite{amc_2023}}, {MinervaMath \cite{lewkowycz2022solving}}, and the \textit{OE\_TO\_maths\_en\_COMP} subset of OlympiadBench \cite{he2024olympiadbench}. The dataset details are introduced in Appendix \ref{app:data}.

\paragraph{Baselines}
We compare \model with the following competitive baselines:

\begin{itemize}[leftmargin=*]
    \item \textbf{Implicit-State Reasoning Models }. These models generate text-only outputs that implicitly maintain all intermediate states. We include three LLMs: \texttt{Qwen3-235B-A22B (Thinking)} \cite{yang2025qwen3} and \texttt{Qwen3-32B (Non-Thinking, Thinking)} \cite{yang2025qwen3}; and three LVLMs: \texttt{GLM-4.5V (108B)}, \texttt{Qwen3-VL-8B-Instruct}, and \texttt{Qwen3-VL-32B-Instruct}.
    \item \textbf{Unconstrained Visual Generation Models}. Unified multimodal models (UMMs) that generate images as part of the reasoning process, including \texttt{Bagel-7B-MoT} \cite{deng2025emerging} and \texttt{Bagel-Zebra-CoT} \cite{li2025zebra}.
    \item \textbf{Tool-Augmented Vision-Language Models (TAVLMs)}. Models that rely on predefined visual tools or APIs operating on given images, including \texttt{DeepEyes} \cite{zheng2025deepeyes} and \texttt{Chain-of-Focus} \cite{zhang2025chain}.
\end{itemize}
We additionally include a text-only RL baseline trained with GRPO on DeepMath-103K, which optimizes trajectory-level correctness and format rewards without executable visual construction or interpreter feedback.
For all baseline models, we set the maximum generation length, sampling temperatures, and other hyperparameters follow the recommended configurations from their respective papers or reports.  The prompt templates are shown in Appendix \ref{subsec:prompt_reasoning}.
Due to space constraints, for UMMs and TAVLMs, we report results on a subset of datasets in the main ablation study for clarity; full results are deferred to Appendix \ref{app:add-exp}.
\begin{table*}[!htbp]
\centering
\small
\setlength{\tabcolsep}{7pt}
\begin{tabular}{lcccccccc}
\toprule
{\textbf{Model}} &
{\textbf{\tabincell{c}{AIME \\ 2024}}} &
{\textbf{\tabincell{c}{AIME \\ 2025}}} &
{\textbf{\tabincell{c}{Beyond \\ AIME}}} &
\bf{\tabincell{c}{MATH \\ 500}} &
\bf{AMC} &
\bf{\tabincell{c}{Minerva \\ Math}} &
\bf{\tabincell{c}{Olymp\\ Bench} } &
\bf{Avg.}\\
\midrule

\rowcolor{mygrey!20}
\multicolumn{9}{l}{\textit{Large Language Models}} \\
Qwen3-235B-A22B (Thinking) &83.80 &80.78 &52.00 &95.40 &93.98 &46.69 &74.04 &75.24\\ 
Qwen3-32B (Non-Thinking)  &31.00 &20.20 &16.00 &88.60 &61.45 &39.34 &52.08  &44.10\\
Qwen3-32B (Thinking) & {81.40} & 72.90 &40.00  & {97.30} &{93.98}  &{45.22}  &{71.96}  &71.82\\
\midrule

\rowcolor{mygreen!20}
\multicolumn{9}{l}{\textit{Large Vision-Language Models}} \\
GLM-4.5V (108B) &76.20 &67.34 &47.00 &96.20 &87.95 &38.60 &70.03   &69.62\\
Qwen3-VL-8B-Instruct &61.46 &46.20 &30.00 &94.20 &81.93 &40.07 &67.21 &60.15\\

Qwen3-VL-32B-Instruct &73.33   & 66.20 &43.00   &93.60   &84.34   &41.18   &69.73    &67.34  \\
\midrule

Text-only RL &73.33 &69.22 &46.00 &94.40 &87.95 &43.38 &70.33  &69.23\\
\rowcolor{myspike!20}
{\model (ours)} &  79.58 & {79.32} & {54.00} & {95.00} & {93.98} & {44.49} & {72.40} &{74.11} \\
\textcolor{gaincolor}{\textbf{Gains}} & \textcolor{gaincolor}{\bf{+6.25}} &
\textcolor{gaincolor}{\bf{+13.12}} &
\textcolor{gaincolor}{\bf{+11.00}} &
\textcolor{gaincolor}{\bf{+1.40}} &
\textcolor{gaincolor}{\bf{+9.64}} &
\textcolor{gaincolor}{\bf{+3.31}} &
\textcolor{gaincolor}{\bf{+2.67}} &
\textcolor{gaincolor}{\bf{+6.77}} \\
\bottomrule
\end{tabular}
\caption{Main results (\%) of \model on seven mathematical reasoning benchmarks compared with several competitive baselines. \textcolor{gaincolor}{\textbf{Green-colored}}
font indicates improvement over the baseline Qwen3-VL-32B-Instruct.
}
\vspace{-4mm}
\label{tab:main}
\end{table*}

\paragraph{Evaluation Metrics}
To ensure a stable evaluation, we default to pass@$k$ evaluation \cite{chen2021evaluating} and report pass@1 metrics. 
Specifically, we generate 64 responses for each question of AIME 2024 and AIME 2025 (i.e., $k=64$), and $k=1$ for the remaining datasets. The pass@1 metric is then computed as:
$\text{pass} @ 1=\frac{1}{k} \sum_{i=1}^k p_i$, where $p_i$ denotes the correctness of the $i$-th response.

\subsection{Main Results}
As shown in Table~\ref{tab:main}, \model achieves substantial gains by learning to integrate executable visual construction into multi-turn reasoning. Across seven mathematical benchmarks, \model achieves an average accuracy of 74.11\%, exceeding the base policy (\texttt{Qwen3-VL-32B-Instruct}) by 6.77\% and the text-only RL baseline by 4.88\%.

Performance gains are particularly pronounced on challenging benchmarks such as AIME 2025 and BeyondAIME, where reasoning requires maintaining global geometric or structural constraints. Notably, \model outperforms both equivalently sized text-only reasoning models (e.g., \texttt{Qwen3-32B Thinking}) and a larger LVLM \texttt{GLM-4.5V} that lacks executable visual construction. These results attribute the gains to learning a policy that selectively constructs controllable, executable visual states as intermediate representations during multi-turn reasoning.

\begin{figure*}[!t]
    \centering
    \includegraphics[width=1.0\linewidth]{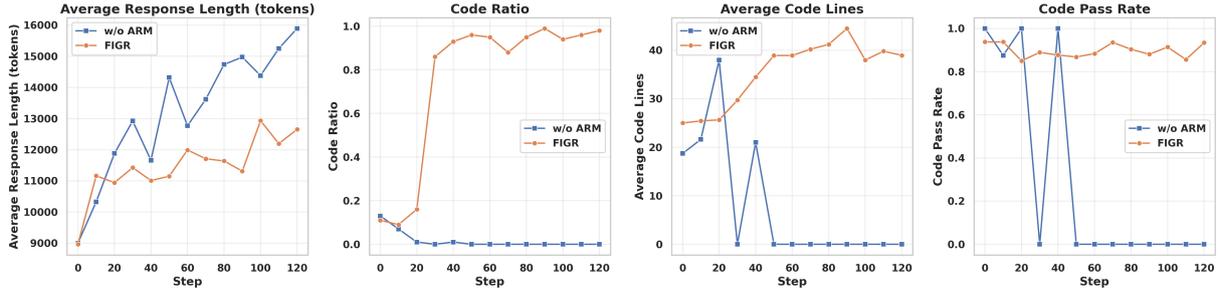}
    \vspace{-6mm}
    \caption{Ablation analysis across training steps.
We track response length, code ratio, code lines, and code pass rate under different settings.
\model exhibits sustained and structured executable visual construction with stable execution behavior, whereas removing the adaptive reward mechanism causes code usage to collapse during training.}
    \label{fig:ablation}
\end{figure*}

\subsection{Ablation Study}

As shown in Table \ref{tab:abla}, we conduct ablation studies on two representative datasets: AIME 2025 and BeyondAIME, to evaluate the contributions of different components in \model.

\paragraph{Prompt Engineering (PE) on the Base Model.} 
Prompt engineering introduces structured multimodal reasoning patterns without updating model parameters. While this leads to moderate performance gains on some datasets, the absence of a learning signal prevents the model from internalizing when and how such behaviors should be applied.
As a result, the induced drawing behaviors remain unstable, which inherently
limits the attainable performance gains.

\begin{table}[!htbp]
\centering
\small
\setlength{\tabcolsep}{12pt}
\begin{tabular}{lcc}
\toprule
\textbf{Model} & \textbf{\tabincell{c}{AIME \\ 2025}} & \textbf{\tabincell{c}{Beyond \\ AIME}} \\
\midrule
Base Model  &66.20 &43.00 \\
\ \ + PE &63.33 &46.00 \\
\ \ + SFT &53.33 &38.00 \\
\ \ + Text-only RL &73.33 &46.00 \\
\ \ + Bagel generated img &64.32 &41.00 \\
\ \ + Qwen generated img &62.93 &41.00  \\
\midrule
Bagel-7B-MoT &10.00 &1.00  \\
DeepEyes &2.34 &0.00  \\
\midrule
\rowcolor{myspike!20}
\textbf{\model (ours)} &  \textbf{79.32} & \textbf{54.00} \\
\ \ w/o ARM &70.00 &49.00 \\
 \ \ w/o visual feedback &76.67 &47.00\\ 

\bottomrule
\end{tabular}
\caption{
Ablation Study on \model. The base model is Qwen3-VL-32B-Instruct, ``PE'' denotes prompt engineering, ``SFT'' denotes supervised fine-tuning, and ``ARM'' denotes the adaptive reward mechanism.
}
\label{tab:abla}
\vspace{-4mm}
\end{table}

\paragraph{Performance Comparison between SFT and RL.}
When fine-tuning the base model on DeepMath-103K, supervised fine-tuning (SFT) leads to a
notable performance drop (e.g., on AIME 2025, accuracy decreases from 66.20\% to 53.33\%). In contrast, text-only reinforcement learning yields consistent performance improvements. This indicates that SFT tends to overfit to training trajectories and generalizes poorly to unseen problems, whereas outcome-driven RL enables more robust and transferable problem-solving behavior.

\paragraph{Injecting Visual Information without Active Visual Reasoning.}
We evaluate two baselines that provide \emph{passively generated diagrams} as static context (\texttt{+Bagel img} and \texttt{+Qwen img}). These diagrams are generated from the problem text using \texttt{Bagel-7B-MoT} and \texttt{Qwen-Image} \cite{wu2025qwen}, respectively, but are not constructed or revised during reasoning.
Although these baselines expose the model to visual content, they do not consistently improve performance over the base model or text-only RL baseline. These results show that passively injected visual inputs, without executable construction or feedback integration, fail to serve as reliable intermediate states. Without a closed reasoning-execution loop, such images neither enforce global constraints nor provide actionable feedback for subsequent reasoning steps.

\paragraph{Performance of UMMs and TAVLMs.} 
Both unified multimodal models and tool-augmented vision-language models exhibit consistently weak performance.
Unified models are vulnerable to cascading errors when tackling highly complex problems.
Tool-augmented methods are restricted to predefined operations on given images and cannot flexibly construct task-specific diagrams.
Consequently, neither paradigm provides reliable and informative visual feedback for complex mathematical reasoning. Due to space limitations, their full results are deferred to Appendix \ref{app:add-exp}.

\paragraph{Ablations on \model Components.}
We further design two ablations to examine the contributions of \model’s components:
(1) Removing the adaptive reward mechanism (ARM). We disable the visual-invocation reward, while keeping all other components unchanged. This ablation removes the explicit control signal that regulates when visual reasoning should be invoked.
(2) Removing visual feedback. We remove only visual feedback during reinforcement learning. The model still performs multi-turn rollouts and executes code, but receives only textual feedback and its corresponding calling reward. This setting isolates the contribution of visual feedback in forming a stable reasoning loop.
Both ablations lead to performance degradation compared to the full \model, demonstrating the effectiveness of the proposed components.

\begin{figure*}[t]
    \centering
    \includegraphics[width=0.95\linewidth]{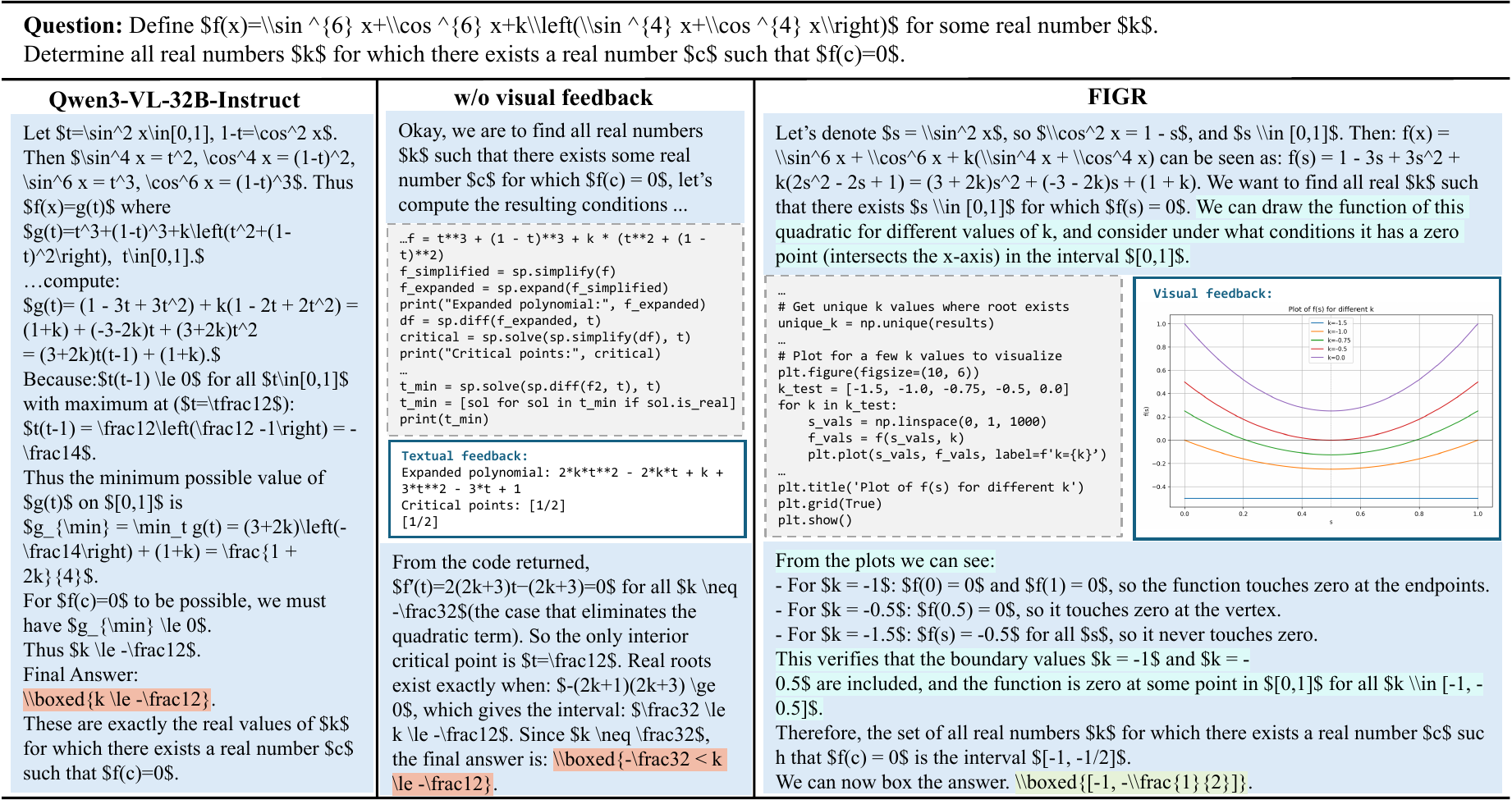}
    \caption{Case study of reasoning behaviors. \model demonstrates more effective integration of visual feedback into the reasoning process, resulting in clearer intermediate reasoning and improved final answers, while the baseline models rely primarily on textual reasoning.}
    \label{fig:case}
    \vspace{-2mm}
\end{figure*}

\paragraph{Evolution of Active Visual Construction During Training.}
We further analyze the emergence of visual-thinking behaviors on the BeyondAIME dataset, using four complementary metrics: response length (measured in tokens with the \texttt{Qwen3-VL-32B-Instruct} tokenizer), code ratio (fraction of samples that invoke code), average code lines (non-empty lines per code block), and code pass rate (fraction of code blocks that execute successfully).

\textbf{(1)} Removing the adaptive reward mechanism (w/o ARM) leads to a brief increase in early code usage. However, this behavior rapidly collapses: both code ratio and code length drop to zero as the model learns that indiscriminate code invocation does not improve outcomes. This pattern reflects an overuse–abandonment cycle in the absence of effective reward guidance.

\textbf{(2)}  In contrast, \model maintains a consistently high code ratio throughout training, accompanied by stable and substantially longer code blocks with a high execution pass rate. 
Executable visual feedback provides an external state that enables explicit validation of global structure, which cannot be reliably enforced through implicit textual representations alone.
Consequently, visual thinking becomes a reliable mechanism for validating the whole reasoning process, rather than an opportunistic or exploratory behavior.

\subsection{Case Study}
Figure \ref{fig:case} compares Qwen3-VL-32B-Instruct, the variant without visual feedback, and \model on a representative example. The baseline model relies  on textual reasoning and symbolic manipulation, requiring careful analytical handling of boundary cases, making it prone to subtle oversights. 
The variant without visual feedback introduces multi-turn rollouts and intermediate code
execution, but receives only textual feedback. While this allows the model to verify intermediate numerical values or symbolic expressions, it provides no access to visual cues that reveal global structural properties of the problem (e.g., the overall shape of a function or the location of its zeros). Consequently, reasoning remains largely local and algebraic, and incorrect assumptions are difficult to detect or correct.

In contrast, \model closes the reasoning loop by incorporating explicit visual state, enabling it
to observe global structural patterns and adjust subsequent reasoning steps accordingly.
This leads to a more robust and interpretable solution. Overall, this case study shows that the advantage of \model does not stem merely from executing intermediate programs, but from
actively integrating visual construction into the reasoning process. Compared with purely
textual reasoning or execution with text-only feedback, active visual thinking provides
complementary information that substantially improves reasoning reliability.
\section{Conclusion}
We introduce \model, which integrates executable visual construction into multi-turn reasoning via end-to-end reinforcement learning. By embedding controllable and revisable visual states within the reasoning loop, \model enables reliable reasoning over global structural constraints that are difficult to maintain through text alone. Extensive experiments demonstrate that learning when and how to externalize intermediate structure as executable visual states significantly improves reasoning stability and accuracy on complex mathematical problems.

\section*{Limitations}
While our results demonstrate the effectiveness of executable visual construction for complex reasoning, several limitations remain that suggest directions for future work.
The proposed method incurs additional computational cost due to multi-turn reasoning and code execution. This overhead is inherent to approaches that externalize intermediate states, motivating future efforts toward more efficient reasoning–execution loops through compact visual representations or adaptive control strategies.
Executable visual construction is particularly beneficial for problems that require maintaining global structural or relational constraints. In contrast, tasks that can be easily solved through local or purely symbolic reasoning may see less benefit from visual externalization. More precise identification and characterization of such task regimes remains an open research question.
Our evaluation focuses on mathematical reasoning benchmarks that emphasize structural consistency and verifiable outcomes. Extending this paradigm to other domains, including scientific reasoning, program synthesis, and long-horizon planning, represents a natural and promising direction for future work.

\bibliography{custom}
\clearpage
\appendix
\begin{table*}[!htbp]
\centering
\small
\setlength{\tabcolsep}{7pt}
\begin{tabular}{lcccccccc}
\toprule
{\textbf{Model}} &
{\textbf{\tabincell{c}{AIME \\ 2024}}} &
{\textbf{\tabincell{c}{AIME \\ 2025}}} &
{\textbf{\tabincell{c}{Beyond \\ AIME}}} &
\bf{\tabincell{c}{MATH \\ 500}} &
\bf{AMC} &
\bf{\tabincell{c}{Minerva \\ Math}} &
\bf{\tabincell{c}{Olymp\\ Bench} } &
\bf{Avg.}\\
\midrule

\rowcolor{mygrey!20}
\multicolumn{9}{l}{\textit{Large Language Models}} \\
Qwen3-235B-A22B (Thinking) &83.80 &80.78 &52.00 &95.40 &93.98 &46.69 &74.04 &75.24\\ 
Qwen3-32B (Non-Thinking)  &31.00 &20.20 &16.00 &88.60 &61.45 &39.34 &52.08  &44.10\\
Qwen3-32B (Thinking) & {81.40} & 72.90 &40.00  & {97.30} &{93.98}  &{45.22}  &{71.96}  &71.82\\
\midrule

\rowcolor{mygreen!20}
\multicolumn{9}{l}{\textit{Large Vision-Language Models}} \\
GLM-4.5V (108B) &76.20 &67.34 &47.00 &96.20 &87.95 &38.60 &70.03   &69.62\\
Qwen3-VL-8B-Instruct &61.46 &46.20 &30.00 &94.20 &81.93 &40.07 &67.21 &60.15\\

Qwen3-VL-32B-Instruct &73.33   & 66.20 &43.00   &93.60   &84.34   &41.18   &69.73    &67.34  \\
\midrule

\rowcolor{myblue!20}
\multicolumn{9}{l}{\textit{Unified Multimodal Models}} \\
Bagel-7B-MoT &10.00 &3.33 &1.00 &56.00 &22.89 &9.56 &26.11  &18.70\\
Bagel-Zebra-CoT (7B) &0.00 &5.55 &0.00 &20.60 &9.64 &2.94 &8.46 & 6.74\\
\midrule
\rowcolor{mypurple!20}
\multicolumn{9}{l}{\textit{Tool-Augmented Vision-Language Models}} \\
DeepEyes &6.98 &2.34 &0.00 &11.00 &18.07 &5.15 &13.65  &8.17 \\
Chain-of-Focus &1.30 &0.00 &0.00 &1.20 &0.00 &0.74 &0.45 &0.53\\
\midrule

Text-only RL &73.33 &69.22 &46.00 &94.40 &87.95 &43.38 &70.33  &69.23\\
\rowcolor{myspike!20}
{\model (ours)} &79.58 & {79.32} & {54.00} & {95.00} & {93.98} & {44.49} & {72.40} &{74.11} \\
\textcolor{gaincolor}{\textbf{Gains}} & \textcolor{gaincolor}{\bf{+6.25}} &
\textcolor{gaincolor}{\bf{+13.12}} &
\textcolor{gaincolor}{\bf{+11.00}} &
\textcolor{gaincolor}{\bf{+1.40}} &
\textcolor{gaincolor}{\bf{+9.64}} &
\textcolor{gaincolor}{\bf{+3.31}} &
\textcolor{gaincolor}{\bf{+2.67}} &
\textcolor{gaincolor}{\bf{+6.77}} \\
\bottomrule
\end{tabular}
\caption{Full experimental results (\%) on seven mathematical reasoning benchmarks compared with several competitive baselines. \textcolor{gaincolor}{\textbf{Green-colored}}
font indicates improvement over the baseline Qwen3-VL-32B-Instruct.
}
\label{tab:full}
\end{table*}

\begin{figure*}[!htbp]
    \centering
    \includegraphics[width=1.0\linewidth]{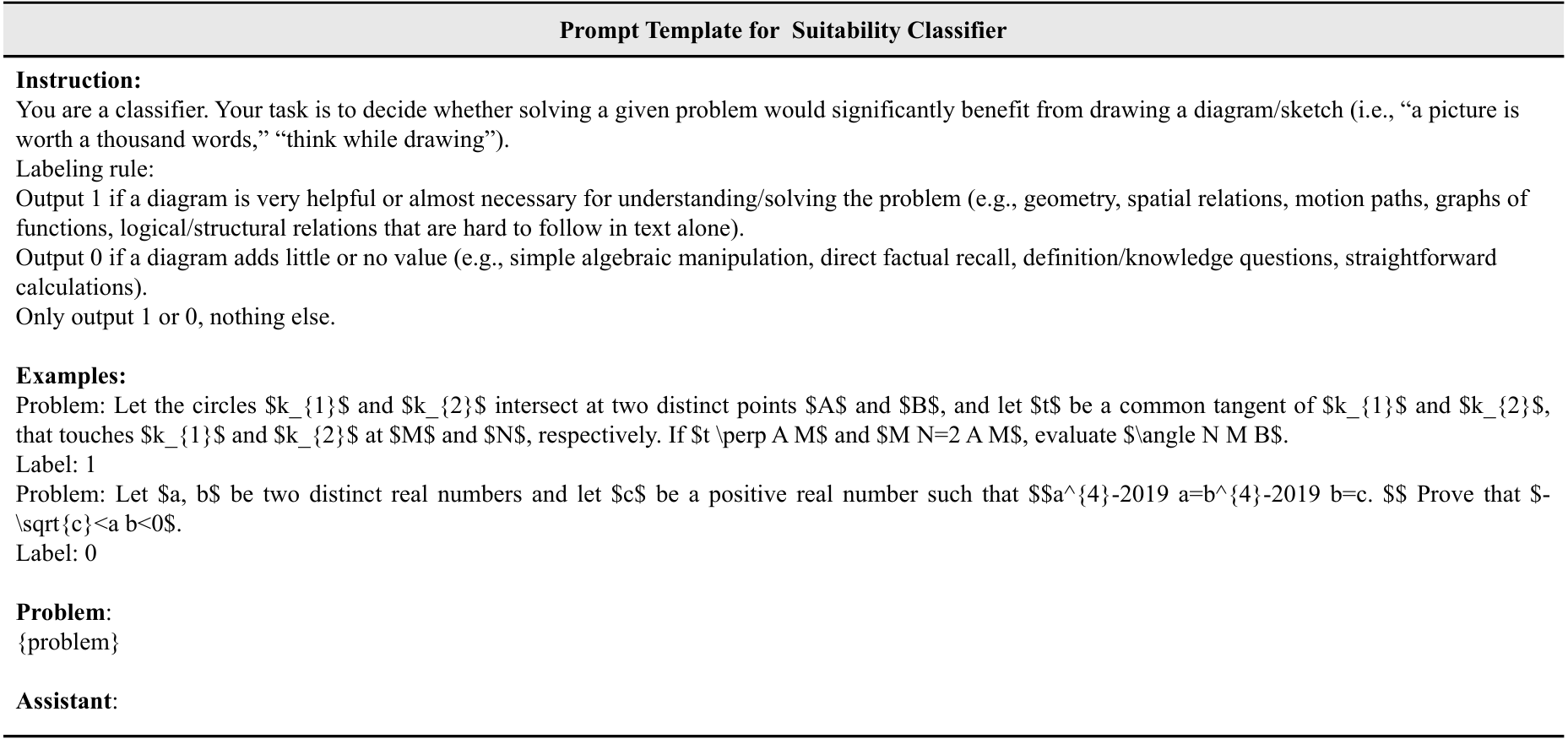}
    \caption{Prompt template for the suitability classifier.}
    \label{fig:prompt0}
\end{figure*}

\section{Dataset Details}
\label{app:data}
We evaluate models on a suite of challenging mathematical reasoning benchmarks, spanning competition problems, standardized mathematics datasets, and advanced reasoning tasks:

(1) \textbf{AIME 2024} \cite{aime_2024} dataset contains 30 problems from the American Invitational Mathematics Examination 2024. Each problem requires multi-step reasoning across algebra, number theory, combinatorics, geometry, and probability.

(2) \textbf{AIME 2025} \cite{aime_2025}. Like the AIME 2024 dataset, the AIME 2025 dataset comprises the 30 problems from the 2025 AIME competitions. 

(3) \textbf{Beyond AIME} \cite{seed2025seed1} is a recently proposed benchmark of 100 problems. It is designed to push beyond standard AIME problems by emphasizing questions that reduce memorization and increase the complexity of reasoning.

(4) \textbf{MATH 500} \cite{hendrycks2021measuring, lightman2023let}. The MATH dataset contains 12,500 problems with step-by-step solutions spanning diverse mathematical topics. For evaluation focused on high-difficulty items, the MATH 500 subset (500 problems) is often used.

(5) \textbf{AMC} \cite{amc_2023}. The American Mathematics Competitions (AMC) dataset consists of 83 problems from the AMC series, a set of standardized mathematics contests administered annually. 

(6) \textbf{MinervaMath} \cite{lewkowycz2022solving} includes 272 undergraduate-level quantitative reasoning problems that require logical deductions and multi-step solutions, serving as a more advanced testbed beyond high-school competition problems.

(7) \textbf{OlympiadBench} \cite{he2024olympiadbench} is an Olympiad-level bilingual multimodal scientific benchmark that includes thousands of challenging mathematics and physics problems compiled from international competitions, with detailed annotations for step-by-step reasoning. In this work, we specifically evaluate the math-focused English comprehensive subsets from OlympiadBench: \emph{OE\_TO\_maths\_en\_COMP}. This subset is an English open-ended, text-only, comprehensive math subset containing 674 problems, emphasizing rigorous mathematical reasoning in text form. 

\section{Prompt Templates}
\label{app:prompt}
\subsection{Prompt Template for Suitability Classifier}
\label{subsec:prompt_classifier}
We provide a dedicated prompt template for the suitability classifier as introduced in Section \ref{subsec:arm} and Section \ref{subsec:setup}, which is used to determine whether a given problem is suitable for diagrammatic  reasoning. As shown in Figure \ref{fig:prompt0}, the template instructs the model to assess the necessity and usefulness of visual sketches based solely on the problem statement.

\subsection{Prompt Template for Reasoning}
\label{subsec:prompt_reasoning}
We adopt three different categories of prompt templates corresponding to the reasoning settings.
(1) As shown in Figure \ref{fig:prompt1}, a multi-turn prompt is used for \model, the variant without visual feedback, and the variant without ARM. Models can iteratively generate output and receive feedback from the interpreter.
(2) As shown in Figure \ref{fig:prompt2}, a standard single-turn prompt is adopted for the text-only RL baseline and other models, requiring the model to directly output the final answer without intermediate interactions.
(3) For DeepEyes and Chain-of-Focus, we adopt the prompt templates recommended by their original papers.

\begin{figure*}[!htbp]
    \centering
    \includegraphics[width=1.0\linewidth]{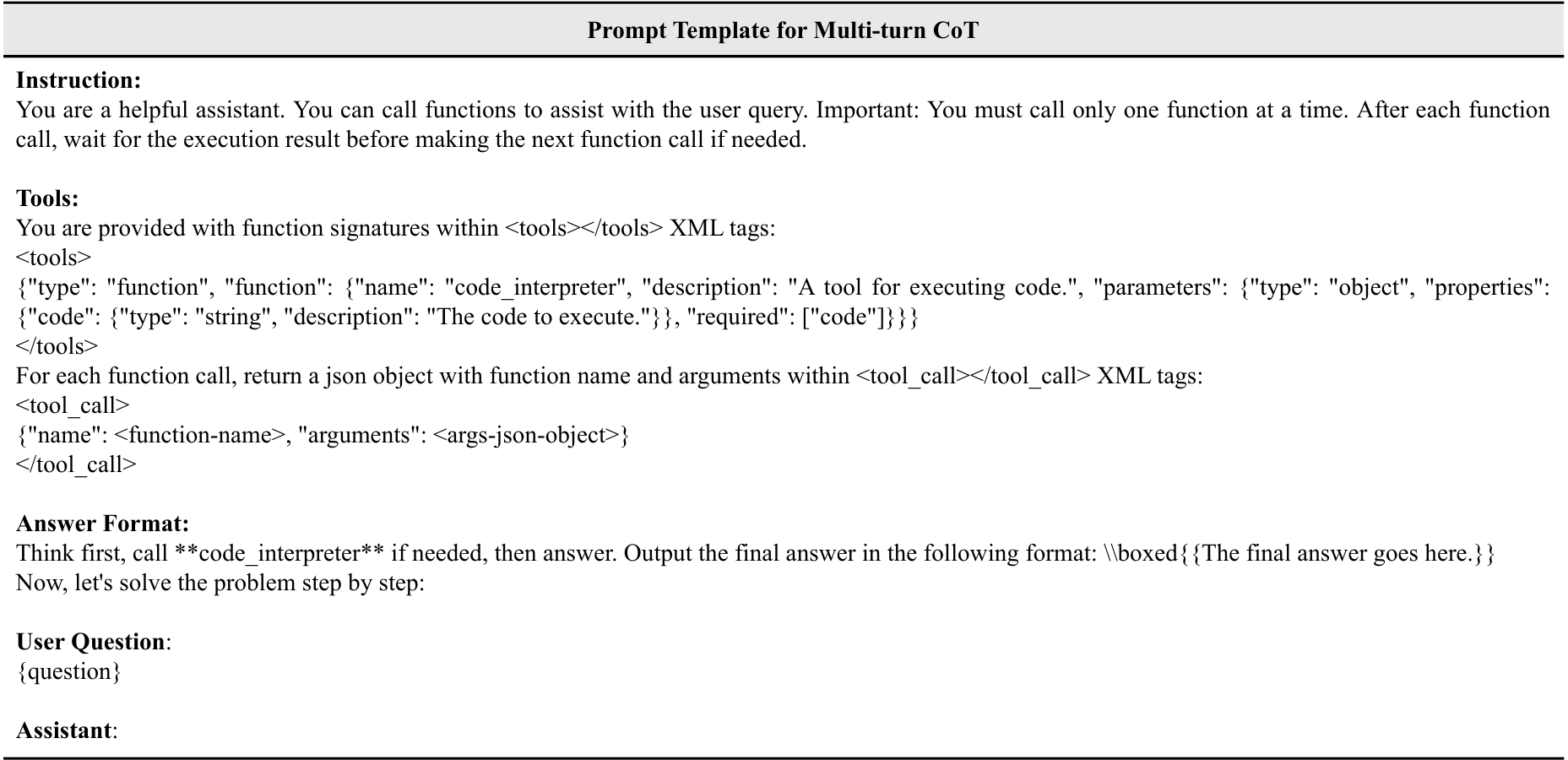}
    \caption{Prompt template for multi-turn CoT.}
    \label{fig:prompt1}
\end{figure*}
\begin{figure*}[!htbp]
    \centering
    \includegraphics[width=1.0\linewidth]{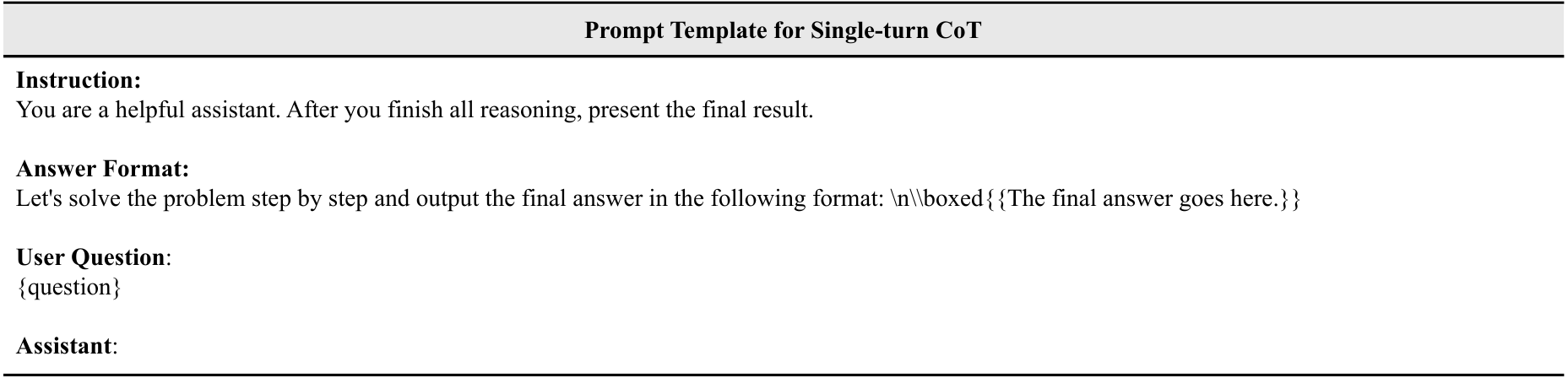}
    \caption{Prompt template for single-turn CoT.}
    \label{fig:prompt2}
\end{figure*}

\section{Additional Experimental Results}
\label{app:add-exp}
As shown in Table \ref{tab:full}, we present the full experimental results, including unified multimodal models and tool-augmented vision-language models that are partially reported in the main context.
\end{document}